\documentclass[letterpaper, 10 pt, conference]{ieeeconf}  

\IEEEoverridecommandlockouts                              

\usepackage{amsmath,amsfonts}
\usepackage{algorithm}
\usepackage{cite}
\usepackage{multirow}
\usepackage{caption}
\usepackage{mathtools}
\usepackage{tabularx}
\usepackage{booktabs}
\usepackage{optidef}
\usepackage{algpseudocode}
\usepackage{array}
\usepackage[caption=false,font=normalsize,labelfont=sf,textfont=sf]{subfig}
\usepackage{textcomp}
\usepackage{color}
\usepackage{url}
\usepackage{verbatim}
\usepackage{graphicx}

\title{\LARGE \bf
Interaction-Aware Vehicle Motion Planning with Collision Avoidance Constraints in Highway Traffic
}

\author{Dongryul Kim, Hyeonjeong Kim and Kyoungseok Han
\thanks{*This research was supported in part by Basic Science Research Program through the National Research Foundation of Korea (NRF), funded by the Ministry of Education (NRF-2021R1A6A1A03043144); in part by the NRF grant, funded by the Korean government (MSIT) (NRF-2021R1C1C1003464); in part by the Technology Innovation Program (20014121, Development of Integrated Minimal Risk
Maneuver Technology for Fallback system during Autonomous Driving) funded by the Ministry of Trade, Industry \& Energy (MOTIE, Korea).}
\thanks{Dongryul Kim, Hyeonjeong kim and Kyoungseok Han (Corresponding Author) are with Department of the Mechanical Engineering, Kyungpook National University, Daegu 41566, South Korea; email:
        {\tt\small ddyul, hyeonjeong9, and kyoungsh @knu.ac.kr.
        }}%
}

\begin{document}

\maketitle
\thispagestyle{empty}
\pagestyle{empty}

\begin{abstract}
This paper proposes collision-free optimal trajectory planning for autonomous vehicles in highway traffic, where vehicles need to deal with the interaction among each other. To address this issue, a novel optimal control framework is suggested, which couples the trajectory of surrounding vehicles with collision avoidance constraints. Additionally, we describe a trajectory optimization technique under state constraints, utilizing a planner based on Pontryagin's Minimum Principle, capable of numerically solving collision avoidance scenarios with surrounding vehicles. Simulation results demonstrate the effectiveness of the proposed approach regarding interaction-based motion planning for different scenarios.
\end{abstract}

\section{Introduction}

With the increasing traffic density, autonomous vehicles have emerged as a promising solution to address issues related to emissions, energy efficiency, and driving safety. Their intelligent capabilities have made them a promising area of research within the autonomous industry \cite{c1,c2,c3}.

Ensuring safety is paramount in the development of autonomous vehicles, as it directly impacts human lives in modern intelligent transport systems. Lane-changing maneuvers are often employed to enhance safety in the presence of obstacles during driving \cite{c4,c5}. Performing such maneuvers requires a sequence of steps, including obstacle recognition, identification of hazardous areas, selection of optimal avoidance maneuvers, and generation and tracking of a collision-free trajectory \cite{c6,c7}. We place emphasis on planning a lane-changing trajectory that ensures safety-critical traffic scenarios. This study considers one such scenario: highway driving maneuvers for collision avoidance with surrounding vehicles. Such maneuvers rely heavily on intricate driver-to-driver interactions, presenting challenges due to limited communication possibilities between vehicles.


Trajectory planning with collision avoidance constraints has to be considered to avoid moving obstacle in highway traffic. Previous research has tackled this challenge using Model Predictive Control (MPC) \cite{c8, c9}. The MPC path planning framework offers a valuable solution to a constrained optimal control problem over a finite time horizon \cite{c10, c11}. In particular, this approach can optimize a cost function considering vehicle dynamics, physical constraints, and collision avoidance constraints. The optimal control problem under constraints is managed through a receding horizon approach, where the problem is iteratively solved at each time step with a horizon that shifts based on updated sensor measurements. This method ensures collision avoidance with surrounding vehicles, assuming the feasibility of the optimization problem. However, when predicting the behavior of surrounding traffic without vehicle-to-vehicle communication, forecasting behavior over the horizon becomes essential for driving safety and traffic interactions. The prediction of vehicle trajectories is critical for capturing interactions among surrounding vehicles, and these predicted trajectories need to be integrated into the collision avoidance constraints of a trajectory planner \cite{c12}.

To address this issue, we propose new approach for coupling the predictor with the trajectory planner, leveraging Pontryagin's minimum principle (PMP) \cite{c13}. This method involves using the predicted trajectories of surrounding vehicles to formulate an optimal control solution based on PMP. Here, state constraints function as collision avoidance constraints, establishing a connection with the predicted trajectory. To solve the optimal control problem, necessary condition and jump condition are introduced, which can provide a novel solution for minimization of acceleration and collision avoidance constraints to secure driving comfort and safety respectively. To show effectiveness of our approach, a novel problem for interaction-dependent traffic is introduced and we demonstrate promising simulation results for various interactive traffic scenarios.

The remainder of the study is organized as follows. In Section~\ref{optimal control}, we provide an introduction to optimal control theory based on PMP. In Section~\ref{problem}, we present the problem of collision avoidance in a surrounding traffic environment, including a description of trajectory predictor, collision avoidance constraints, and how they are incorporated into the optimization problem. In this section, we also introduce a optimal trajectory planning method based on necessary and jump conditions for optimization. In Section~\ref{simulation}, we verify the effectiveness of our approach using simulations in various environments and analyze the simulations for different scenarios. Finally, in Section~\ref{conclusion}, we conclude the study by summarizing the key findings.
\section{Optimal Control}\label{optimal control}

\subsection{Optimal Control Problem}

A controlled vehicle system is described by a set of ordinary
differential equations that show how the system changes over time. When incorporating optimal control principles into the system, the objective is to optimize a performance index while satisfying boundary conditions, state equations, and state constraints. Thus, we frame the optimal control problem as follows \cite{c14}:
\begin{mini!}
{u^*}{J =\int_{t_{0}}^{t_{f}}L(u(t),t)dt, \label{COSTFUNCTION} } 
{}{}
\addConstraint{\dot{x}(t)=f(x(t),u(t),t) \label{STATE EQ} }
\addConstraint{x(t_0) = x_{0}, x(t_f) = x_{f}  \label{boundary condition} }
\addConstraint{h \left(x(t_{d}),t_{d}\right)=0 \label{coupling}}
\end{mini!}
where $J$ is cost functional or performance index, the integrand function $L(\cdot)$ is the Lagrange performance index, $f(\cdot)$ is the vector of state equation function which consists of $x(t)$ and $u(t)$, $t$ represents the time variable which is defined from $t_0$ to $t_f$, $t_0$ and $t_f$  stands for initial and final time, $x_0$ and $x_f$ stands for initial and final state value, $u(t)$ is the vector of control variable, $x(t)$ is the vector of state variables, $h(\cdot)$ is state-equality constraints, which is a function of state and time. An optimal trajectory can be created by minimizing the cost functional $J$ while meeting the boundary condition \eqref{boundary condition} and state-equality constraints \eqref{coupling}.

\subsection{Pontryagin Minimum Principle}

By utilizing the PMP, the suggested optimal control problems can be resolved through analytical means. So, in this section, necessary and jump condition are introduced to solve the optimal control problem based on PMP. 

First, given \eqref{COSTFUNCTION} and \eqref{STATE EQ}, the Hamiltonian can be defined as:
\begin{equation}
H=L + p \cdot f \label{H}
\end{equation}
where $p$ is the vector of co-state variables which are associated with $f$.

Here, PMP-based necessary condition deal with state equation, co-state equation and optimal control. 

State equations are described as:
\begin{equation} 
\dot{x}(t)=f(x(t),u(t),t)\label{state equation}
\end{equation}

Co-state equations are described as:
\begin{equation} 
\dot{p}(t)=\frac{\partial H( x(t),u(t),p(t),t)}{\partial x} \label{Auxiliary}
\end{equation}

Additionally, it is required that the Hamiltonian along the optimal trajectory, which corresponds to the optimal control $u^*(t)$ and optimal state $x^*(t)$, meets a certain condition:
\begin{equation}
H(x^{*}(t),u^{*}(t),p^{*}(t),t) \le H(x^{*}(t),u(t),p^{*}(t),t).
\end{equation}

Thus, by differentiating $H$ with respect to $u$, we can determine the optimal control as follows: 

\begin{equation}
\frac{ \partial H(x^{*}(t),u^{*}(t),p^{*}(t),t)}{ \partial u}=0 \label{H- minimizing}
\end{equation}

\begin{figure} [t!]
\begin{center}
    \includegraphics[width=85mm]{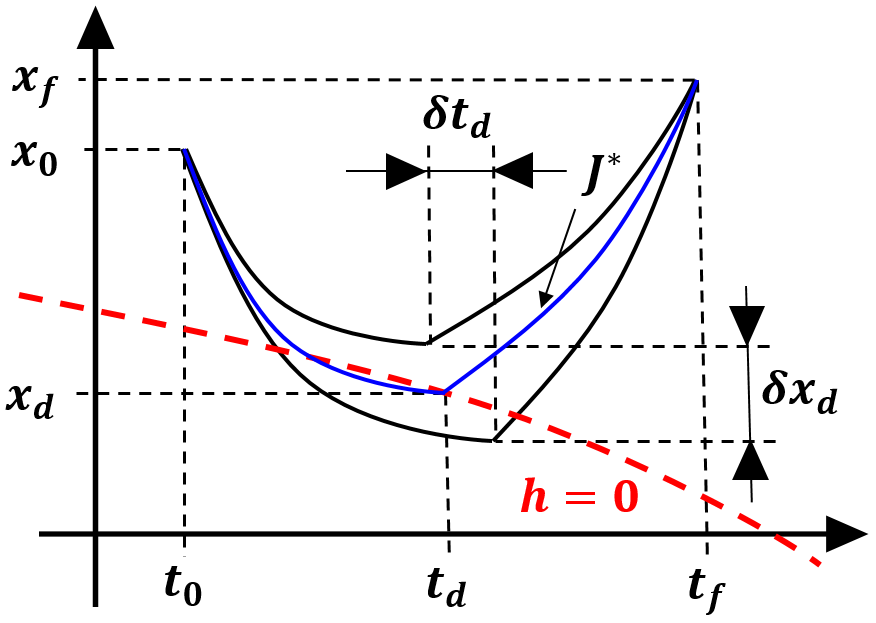}
    \caption{Optimal trajectory with state-equality constraints.}\label{constraints fig}
\end{center}
\end{figure}

Generally, necessary condition is enough to solve the optimal control problem when ignoring the state constraints. However, if a system is specified with state constraints, solving the control problem can become challenging, and it is crucial to handle the application of control at the state constraints with accuracy. Here, we introduce PMP-based state-equality constraints to tackle the control problem, which is used to combine the predicted trajectory and collision avoidance constraints of optimal control in this study.

State-equality constraints can be described as:
\begin{equation}
h \left(x(t_{d}),t_{d}\right)=0
\end{equation}
which means that safety-critical distance $x_d$ can be set up on vehicle trajectory at intermediate time $t_d$, as shown in the Fig.~\ref{constraints fig}. To solve the state-constrained optimal control problem, jump condition is considered: 
\begin{equation}
\lim_{t \to t_{d}^{-}} p (t) = \lim_{t \to t_{d}^{+}} p (t) + \left. \frac{\partial  h\left( x,t_{d} \right )}{\partial x}\right \vert_{x=x(t_{i})} \cdot \Phi
\label{costate dis}
\end{equation}
\begin{equation}
\lim_{t \to t_{d}^{-}} H (t) = \lim_{t \to t_{d}^{+}} H (t) - \left.\frac{\partial h(x^{*}(t_{d}),t)}{\partial t}\right \vert_{t=t_{d}}  
\cdot \Phi
\label{H dis}
\end{equation}
where $\Phi$ is a vector that has the same size as $h$. These equations \eqref{costate dis} and \eqref{H dis} indicate that the $H(\cdot)$ and $p(\cdot)$ can be discontinuous at certain times according to type of functions $h(\cdot)$.  

\section{Modeling and Problem Definition}\label{problem}
\begin{figure} [t!]
\begin{center}
    \includegraphics[width=85mm]{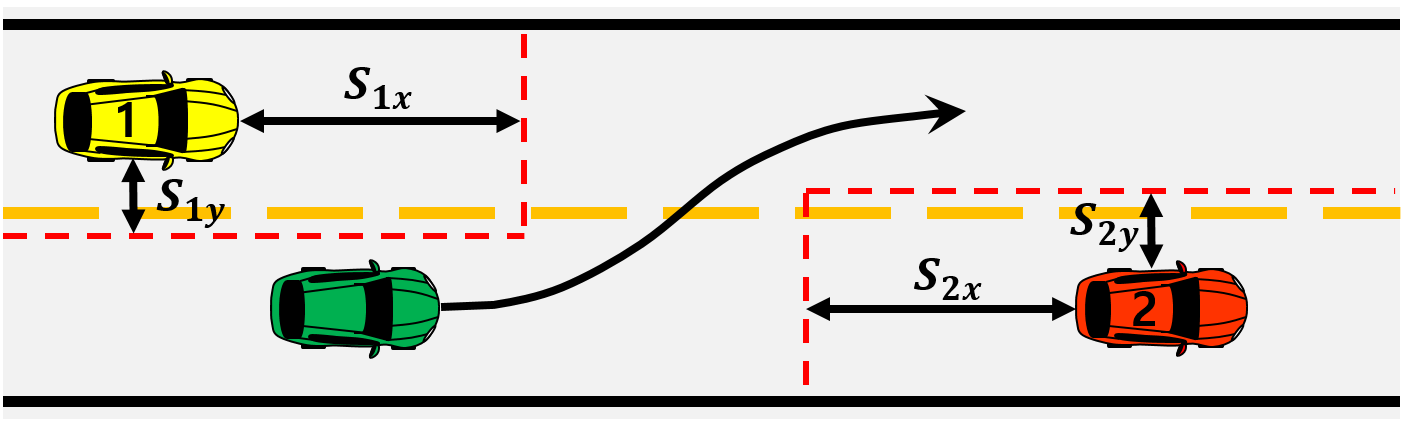}
    \caption{Collision avoidance maneuver during lane change on a road with two lanes. The ego vehicle is depicted in green, the $1$-surrounding vehicle in yellow, and the $2$-surrounding vehicle in red. The red boxes surrounding the vehicles indicate safety-critical regions.}\label{diagram}
\end{center}
\end{figure}

As shown in Fig.~\ref{diagram}, ego vehicle has to plan motion trajectory with objective of avoiding collision with $i$-th surrounding vehicle.
The motion planning approach proposed in this paper aims to avoid safety critical region of surrounding vehicle by creating an optimal trajectory with collision avoidance constraints. This section explains how to plan a trajectory to avoid collision considering the predicted trajectory of surrounding vehicle and interaction among vehicles. 

\subsection{System Dynamics}
We employ a point mass model with 2 degrees of freedom (2DOF) to represent the vehicle's motion in the x and y directions. So, the ego-vehicle motion can be expressed as:
\begin{equation}
\dot{x}_e = \begin{bmatrix} \dot{s_x} \\ \dot{v_x} \\ \dot{s_y} \\ \dot{v_y} \end{bmatrix} = \begin{bmatrix} {v_x} \\ {u_x} \\ {v_y}\\ {u_y}\end{bmatrix}, \; u_e = \begin{bmatrix} {u_x} \\ {u_y} \end{bmatrix}
\end{equation}

where $x_e$ is the state, including the $x-$ and $y-$ motion of the ego-vehicle, $u_e$ is the control input, encompassing the $x-$ and $y-$ acceleration of the ego-vehicle, $s_x$ and $s_y$ are the longitudinal and lateral displacements, and $v_x$ and $v_y$ denote the longitudinal and lateral velocities. Additionally, $u_x$ and $u_y$ represent the longitudinal and lateral accelerations, respectively.

So, this dynamics model expresses the general state equation as: 
\begin{equation} 
\dot{x}_e(t)=f_e(x_e(t),u_e(t),t)\label{state equation}
\end{equation}

\subsection{Prediction using Markov Chain}
In this study, it is preacquisition to predict the trajactory of surrounding vehicle before creating an optimal trajectory. The Markov chain is the proposed algorithm which is essential to predict location of surrounding vehicles and is based on corrected data. The first-order Markov chain has a strength that it assumes future state only depends on present state. The transition probability matrix can be represented as:
\begin{equation} \label{markov}
  \begin{split}
    P(x_{t+1}=\bar{x}_j|x_{t}=\bar{x}_k,x_{t-2}=\bar{x}_{k-1},...,x_0=\bar{x}_0) \\
      = P(x_{t+1}=\bar{x}_{j}|x_t=\bar{x}_k)
  \end{split}
\end{equation}
where $x_{t}$ is the position probability distribution of surrounding vehicle at time $t$ and $j$, $k$ denote the indices of the states in the state space. In this case the prediction has lower accuracy than high-order ($n$th-order) Markov chain.

High-order Markov chain uses not only present state but past state. However the conventional model for a $n$th-order Markov chain has $(m-1)m^n$ model parameters, where $m$ is number of state. This causes huge computational time why a high-order Markov chain is unusable in practical region. To overcome disadvantage, a modified high-order Markov chain which includes only on additional parameter for each lag $\lambda$ was proposed by Raftery\cite{c15}. The transition probability matrix could be rewritten as follows:
\begin{equation}
    \begin{split}
  P(x_{t+1}=\bar{x}_j|x_{t}=\bar{x}_1,x_{t-1}=\bar{x}_{2},...,x_{t-n+1}=\bar{x}_{n}) \\
  =\sum\limits_{i=1}^n \lambda_iq_{\bar{x}_j\bar{x}_i}\label{markov2}
    \end{split}
\end{equation}
and
\begin{subequations}\label{constraint}
\begin{align}
         \sum\limits_{i=1}^n \lambda_i = 1\label{constraint1}\\
        0\leq\sum\limits_{i=1}^n \lambda_iq_{\bar{x}_j\bar{x}_i}\leq1 \label{constraint2}    
\end{align}
\end{subequations}

$\lambda_i$ is the lag parameters which are non-negative real numbers, $q_{\bar{x}_j\bar{x}_i}$ is the probability of a transition matrix $Q$ and $n$ is order of Markov chain. $Q$ matrix differs depending on each lag $n$. With \eqref{markov2}, \eqref{constraint1}, and \eqref{constraint2}, discrete prediction data of surrounding vehicles is obtained as:
\begin{equation}\label{prediction}
  \hat{x}_{t}=\sum\limits_{i=1}^n \lambda_iQx_{t-i}
\end{equation}

By computing polynomial regression based on the prediction results, we can obtain the trajectory of surrounding vehicle $i$ denoted as $\hat{X}_{i}$. This trajectory can then be integrated with collision avoidance constraints in the time domain. Now, introducing a predictor denoted as $\Omega$, we can anticipate the future trajectory $\hat{X}_{i}$. This predictor utilizes past trajectories of surrounding vehicles $(X_{obs})$ and can be represented as follows:
\begin{equation}\label{prediction_denotion}
  \hat{X}_{i} = \Omega (X_{obs})
\end{equation}

\subsection{Collision Avoidance Constraints}
Once the trajectory of the surrounding vehicle is determined as a polynomial expression, the trajectory $\hat{X}_{i}$ becomes applicable in the optimal control problem. We make the assumption that the $i$-th surrounding vehicle follows a longitudinal trajectory, keeping the lateral position constant. Consequently, the longitudinal position of surrounding vehicle $i$ can be established as a polynomial function derived from $\hat{X}_{i}$, as outlined below:
\begin{equation}\label{obstalce function}
     x_{s, i}(t) = a_i t^3 + b_i t^2 + c_i t + d_i
\end{equation}

\begin{figure}[t!]
    \centering
    \includegraphics[width = 85mm]{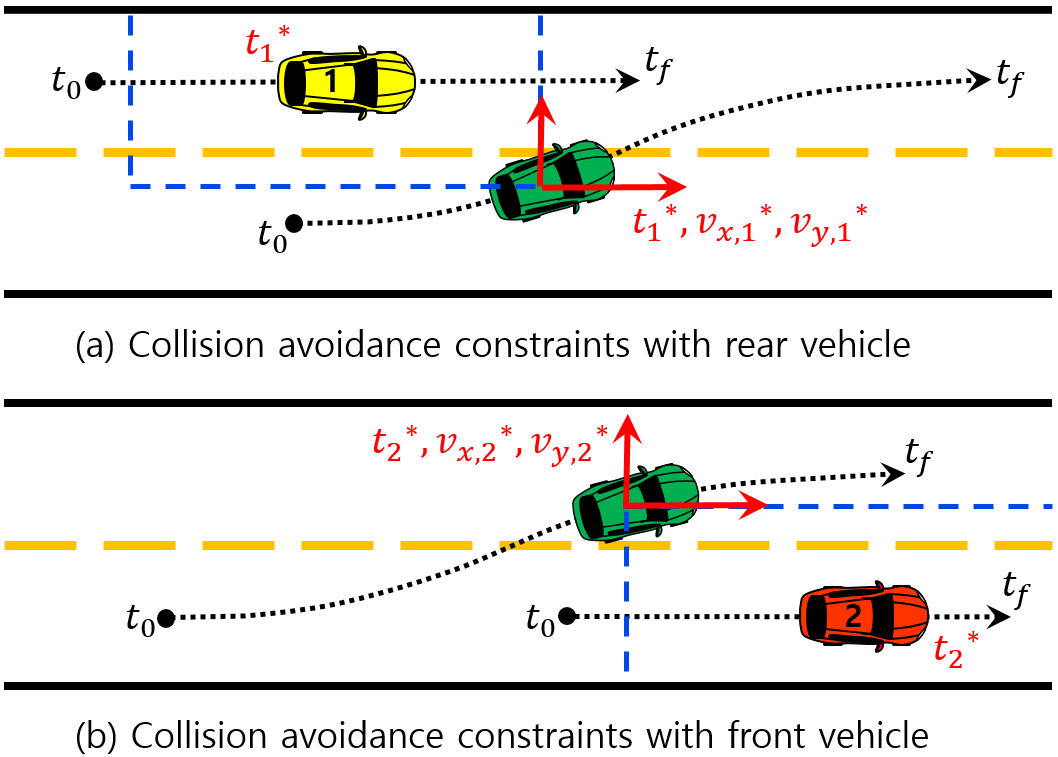}
    \caption{Optimal trajectory planning with collision avoidance constraints, coupled with the predicted trajectory of surrounding vehicles.}\label{collision_diagram}
\end{figure}

Fig.~\ref{collision_diagram} illustrates a principle in which the optimal trajectory is generated by utilizing the edge of the critical safety region of each surrounding vehicle \cite{c16}. In the scenario of lane changing depicted in Fig.~\ref{collision_diagram} (a), the collision avoidance constraint needs to incorporate the predicted trajectory of the $1$-th surrounding vehicle as follows:
\begin{equation}\label{collision1}
    \begin{bmatrix}
s_x(t_1) - ( x_{s, 1} (t_1) + L_{1}/{2} + S_{x1} ) \\
s_y(t_1) - \left( y_{s, 1} - W_{1}/{2} - S_{y1} \right)
\end{bmatrix} = 0
\end{equation}

Likewise, the collision avoidance constraint for the $2$-nd surrounding vehicle in the scenario depicted in Fig.~\ref{collision_diagram} (b) can be formulated as follows:
\begin{equation}\label{collision2}
    \begin{bmatrix}
s_x(t_2) - ( x_{s, 2} (t_2) - L_{2}/{2} - S_{x2} ) \\
s_y(t_2) - \left(  y_{s, 2} + W_{2}/{2} + S_{y2} \right)
\end{bmatrix} = 0
\end{equation}
where $W_{i}$ and $L_{i}$ represent the width and length of the $i$-th surrounding vehicle, respectively, $S_{xi}$ and $S_{yi}$ denote critical-safety distance in the $x$ and $y$ direction from the $i$-th surrounding vehicle, $t_i$ signifies the instant time which is free value allocated for collision avoidance constraints,  $y_{s, i}$ is lateral position of surrounding vehicle $i$.

Here, we can express the constraints \eqref{collision1}, \eqref{collision2} for ego-vehicle and surrounding vehicle $i$ as an equality-state constraints: 
\begin{equation}\label{general collision}
   h_{i}(x_e(t_i), \hat{X}_{i} (t_i), t_i) = 0
\end{equation}
where the predicted trajectory of surrounding vehicle $i$ can be coupled with optimal control problem.

\subsection{Problem Formulation}\label{collision}
In this subsection, we propose an optimal control formulation to obtain an optimal trajectory during lane changes. The objective is to minimize the magnitude and variation of the vehicle's acceleration while considering collision avoidance constraints to enhance driving comfort. Thus, we describe the performance index used to optimize the vehicle trajectory as follows:
\begin{equation}
    L_e(u_e(t),t) = \frac{{a_e}^2}{2} = \frac{(u_x)^2 + (u_y)^2}{2}
\end{equation}
where $a_e$ is the acceleration of the ego vehicle, described by $u_x$ and $u_y$.

To generate the optimal trajectory for lane-changing, $L_e$ has to be optimized while satisfying boundary conditions:
\begin{equation}
 x_e(t_0) =  x_{e,0} = \begin{bmatrix} {s_{x0}} \\ {v_{x0}} \\ {s_{y0}} \\ {0} \end{bmatrix}, \text{ } x_e(t_f) =  x_{e,f}  = \begin{bmatrix} {s_{xf}} \\ {v_{xf}} \\ {s_{yf}} \\ {0} \end{bmatrix} \label{boundary condition2}
\end{equation}
where $s_{x0}$ and $s_{xf}$ represent the initial and final longitudinal positions, $v_{x0}$ and $v_{xf}$ denote the initial and final longitudinal velocities, and $s_{y0}$ and $s_{yf}$ correspond to the initial and final lateral positions, respectively.

Here, we can formulate the optimal control problem for each surrounding vehicle $i$ as:
\begin{mini!}
{{u_e}^*,{x_e}^*, {t_i}^*,} {\int_{t_{0}}^{t_{f}} L_e(u_e(t),t) dt  \label{COSTFUNCTION2}}
{}{}
\addConstraint{\dot{x}_e(t)=f_e(x_e(t),u_e(t),t)}
\addConstraint{
x_e(t_0) =  x_{e,0}, \text{ } x_e(t_f) =  x_{e,f} }
\addConstraint{
h_{i}(x_e(t_i), \hat{X}_{i} (t_i), t_i) = 0}
\addConstraint{\hat{X}_{i} = \Omega (X_{obs})}
\end{mini!}

Based on the optimal control formulation, we have to optimize $t_i^*$, $v_{x,i}^*$ and $v_{y,i}^*$ to create the optimal trajectory.

\subsection{Trajectory Planning using PMP}

In this subsection, we employ the PMP to tackle the optimal control problem discussed in the preceding subsection. The initial step in applying PMP involves formulating the Hamiltonian function. Now, the Hamiltonian function is defined as follows:
\begin{equation} \label{H_3}
  \mathbf{H} = \frac{(u_x)^2 + (u_y)^2}{2} + p_{1}v_{x} + p_{2}u_{x} + p_{3}v_{y} + p_{4}u_{y} 
\end{equation}


To establish the connection between state and co-state variables, the co-state equation is expressed as:
\begin{equation} \label{co-state-eq}
    \begin{bmatrix}
        {\dot p}_1 \\ {\dot p}_2 \\ {\dot p}_3 \\ {\dot p}_4
    \end{bmatrix} = \begin{bmatrix}
        - \mathbf{H}_{s_x} \\ - \mathbf{H}_{v_x} \\ - \mathbf{H}_{s_y} \\ - \mathbf{H}_{v_y} \end{bmatrix}
        = \begin{bmatrix}
        0 \\ - p_1 \\ 0 \\ - p_3
    \end{bmatrix} 
\end{equation}

In minimizing the Hamiltonian, the gradient of $\mathbf{H}$ with respect to the control input are derived as:
\begin{subequations} 
    \begin{align}
        \mathbf{H}_{u_{x}} &= u_x + p_2  \\
        \mathbf{H}_{u_{y}} &= u_y + p_4 
    \end{align}
\end{subequations}

Setting these derivatives to zero yields the optimal values for the control inputs, minimizing the Hamiltonian:
\begin{subequations}\label{optimal}
\begin{align}
    u^{*}_x = - p_2 \\
    u^{*}_y = - p_4 
\end{align} 
\end{subequations}

To address the optimal control problem with collision avoidance constraints \eqref{collision1} or \eqref{collision2}, additional conditions for surrounding vehicles $1$, $2$ are described as:
\begin{subequations}\label{collision condition}
\begin{align}
&\begin{bmatrix}
s_x(t_1) \\
s_y(t_1)
\end{bmatrix} = \begin{bmatrix}
x_{s, 1} (t_1) + L_{1}/{2} + S_{x1} \\
y_{s, 1} - W_{1}/{2} - S_{y1} 
\end{bmatrix} \\
    &\begin{bmatrix}
s_x(t_2) \\
s_y(t_2)
\end{bmatrix} = \begin{bmatrix}
x_{s, 2} (t_2) - L_{2}/{2} - S_{x2}  \\
y_{s, 2} + W_{2}/{2} + S_{y2}
\end{bmatrix}
\end{align}
\end{subequations}
For co-state continuity, jump condition is imposed at $t_i$ as follows:
\begin{equation} \label{jump_costate}
\begin{bmatrix} p_1^{-}(t_i) \\ p_2^{-}(t_i) \\ p_3^{-}(t_i) \\ p_4^{-}(t_i)
\end{bmatrix}
= 
\begin{bmatrix} p_1^{+}(t_i) + \Phi_{i,1}\\ p_2^{+}(t_i) \\ p_3^{+}(t_i) +\Phi_{i,3}\\ p_4^{+}(t_i)
\end{bmatrix}
\end{equation}
where $\Phi_{i,1}$ and $\Phi_{i,3}$ are jump parameters, which can make $p_1$ and $p_3$ discontinuous at $t_i$, respectively. In contrast, $p_2$ and $p_4$ are continuous at $t_i$.

For hamiltonian, the continuity condition can be expressed as: 
\begin{equation} \label{jump_H}
    H^-(t_i) = H^+(t_i) + \dot{x}_{s, i} \Phi_{i,1}
\end{equation}
where continuity of hamiltonian depends on the predicted trajectory of surrounding vehicle $i$. If surrounding vehicle $i$ is static, hamiltonian can be continuous at $t_i$. In contrast, If surrounding vehicle $i$ is moving, hamiltonian can be discontinuous at $t_i$.

Combining the conditions \eqref{boundary condition2}, \eqref{co-state-eq}, \eqref{optimal}, \eqref{collision condition}, \eqref{jump_costate} and \eqref{jump_H}, optimal state and input denoted by $s_{x}^{*}$, $v_{x}^{*}$, $u_{x}^{*}, s_{y}^{*}$, $v_{y}^{*}$, and $u_{y}^{*}$, and $t^*_i$ can be computed. Using the above optimal solution, the ego vehicle can generate optimal trajectory against diverse movements of surrounding vehicle in lane change, which achieve to couple the collision avoidance constraints and the predicted trajectory.  

\section{Simulation Result}\label{simulation}
The simulation of autonomous driving on a two-lane road is conducted using the Matlab environment. Three sample scenarios are explored, wherein an optimal trajectory is generated while considering collision avoidance constraints with the rear surrounding vehicle 1, as illustrated in Fig.~\ref{collision_diagram}(a), situated in the adjacent lane.

Several parameter settings are employed to simulate the optimal trajectory. Firstly, the origin of the $x, y$-coordinates has to be set. So, initial longitudinal position of the ego set as $0$, with $s_{x0} = 0[m]$, $s_{y0} = -2[m]$ in \eqref{boundary condition2}. The velocity of the ego vehicle is assumed to remain constant, with $v_{x0} = v_{xf} = v_e$. The width of the road, represented by $s_{yf}$, is set to $3.75[m]$. Table~\ref{table} provides details on different parameters for the three scenarios, specifying values for $v_e$, $s_{xf}$, and $t_f$ for the ego vehicle, as well as initial longitudinal position and velocity of surrounding vehicle $i$. Based on these settings, the trajectory planning performance is evaluated for each of the three scenarios.

\begin{table}[h!]
\centering
{\scriptsize
\setlength{\tabcolsep}{5pt}
\renewcommand{\arraystretch}{1.5}

\caption{Scenarios setting.} \label{table}

\begin{tabular}{c c c c c c} 
 \hline  
\scriptsize{Scenario} & \scriptsize{$v_{e}[m/s]$} & \scriptsize{$s_{xf}[m]$} & \footnotesize{$t_{f}[s]$}  & \footnotesize{$x_{s, i}(t_0)$[m]} & \footnotesize{$v_{s, i}(t_0)$[m/s]} \\\hline \hline  
1 & 20 & 85 & 5 & -15 & 18\\\hline
2 & 20 & 73 & 5 & -15 & 20\\\hline
3 & 20 & 90 & 5 & -15 & 13
  \\ \hline  
\end{tabular}}

\end{table}%

In the scenario 1, a lane-changing collision avoidance maneuver is executed when surrounding vehicle 1 continues driving at a specific speed before $t_0$ in the adjacent lane. The scenario 2 simulates the optimal trajectory when surrounding vehicle 1 carries out a deceleration maneuver before $t_0$. In the scenario 3, surrounding vehicle 1 accelerates before $t_0$. In all three scenarios, the ego vehicle maintains its speed and generates an optimized trajectory that is coupled with the trajectory of surrounding vehicle 1. In the simulations, the predictor assesses the situation based on the estimated trajectory of the surrounding vehicle, and the trajectory planner determines the appropriate lane-changing maneuver to prevent collisions with three cases of surrounding vehicle $1$.

For the three scenarios, the same speed and lane-change duration were set for the ego vehicle, and identical initial positions were set for the surrounding vehicles. However, depending on the behavior of the surrounding vehicles, it is necessary to ensure a safety distance at the final time. Therefore, in case the adjacent vehicle accelerates and drives aggressively, $s_{xf}$ is set to a larger value, while in the scenario where the adjacent vehicle brakes with a more conservative driving style, $s_{xf}$ is set to a smaller value.

\begin{figure} [t!]
\begin{center}
    \includegraphics[width=85mm]{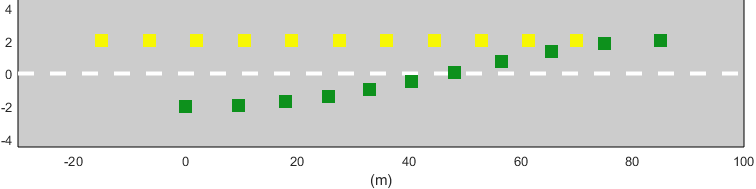}
    \caption{The position of the ego vehicle and surrounding vehicle 1 is recorded every 0.5 seconds, denoted by the $(x, y)$ coordinates, where surrounding vehicle 1 travels at a constant velocity on the adjacent road (scenario 1). }\label{scenario 1}
\end{center}
\end{figure}

\begin{figure} [t!]
\begin{center}
    \includegraphics[width=85mm]{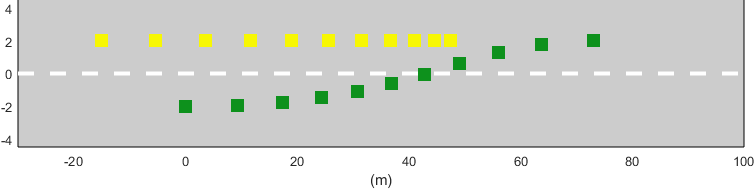}
    \caption{The position of the ego vehicle and surrounding vehicle 1 is recorded every 0.5 seconds, denoted by the $(x, y)$ coordinates, where surrounding vehicle 1 decelerates on the adjacent road (scenario 2).}\label{scenario 2}
\end{center}
\end{figure}

\begin{figure} [t!]
\begin{center}
    \includegraphics[width=85mm]{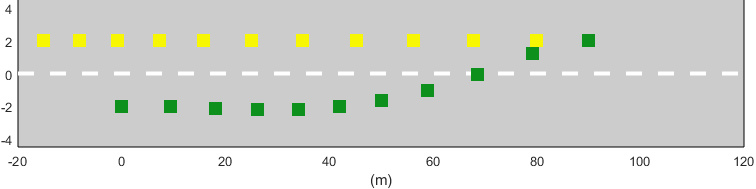}
    \caption{The position of the ego vehicle and surrounding vehicle 1 is recorded every 0.5 seconds, denoted by the $(x, y)$ coordinates, where surrounding vehicle 1 accelerate on the adjacent road (scenario 3).}\label{scenario 3}
\end{center}
\end{figure}

\begin{figure} [t!]
\begin{center}
    \includegraphics[width=85mm]{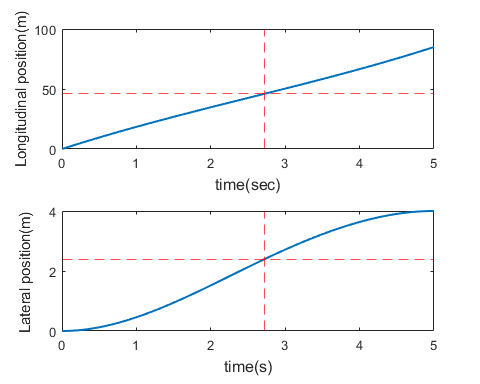}
    \caption{Simulation of optimal longitudinal and lateral position trajectory (scenario 1).}\label{constant_vel}
\end{center}
\end{figure}

\begin{figure} [t!]
\begin{center}
    \includegraphics[width=85mm]{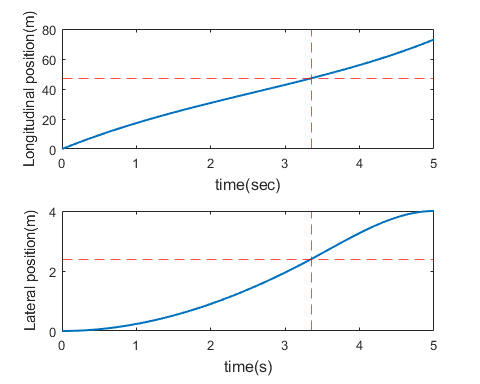}
    \caption{Simulation of optimal longitudinal and lateral position trajectory (scenario 2).}\label{brake}
\end{center}
\end{figure}

\begin{figure} [t!]
\begin{center}
    \includegraphics[width=85mm]{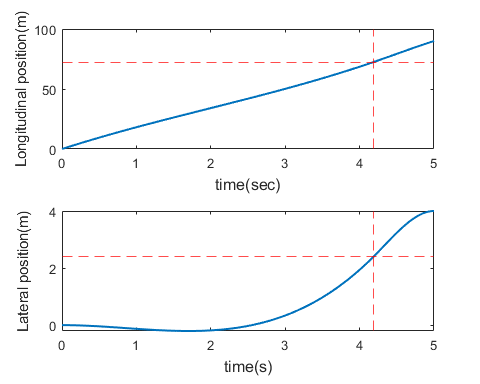}
    \caption{Simulation of optimal longitudinal and lateral position trajectory (scenario 3).}\label{acc}
\end{center}
\end{figure}

Figs.~\ref{scenario 1}, \ref{scenario 2}, and \ref{scenario 3} present simulation results based on the motion of the surrounding vehicle in each scenario. These figures allow us to observe interaction tendencies corresponding to the dynamic behavior of the surrounding vehicle. In Fig.~\ref{scenario 1}, the ego vehicle generates a collision-free trajectory in response to the constant speed of the surrounding vehicle. When the surrounding vehicle decelerates, the ego vehicle plans a more aggressive behavior as shown in the Fig.~\ref{scenario 2}. Conversely, when the surrounding vehicle accelerates aggressively, the ego vehicle exhibits a more conservative behavior compared to scenario 1, as shown in the Fig.~\ref{scenario 3}. This validates our approach, demonstrating that the trajectory planner can generate an optimized trajectory in which the predicted trajectories for each case can be effectively coupled. Figs.~\ref{constant_vel}, \ref{brake}, and \ref{acc} illustrate the position trajectories over time. These figures depict collision avoidance constraints by considering the safety-critical region at $t^*_1$, represented by red dashed lines on each position graph.

\section{Conclusion}\label{conclusion}
This paper introduces an advanced collision-free optimal trajectory planning methodology tailored for autonomous vehicles operating in highway traffic, demanding effective management of interactions among vehicles. To address this complex challenge, we present an innovative optimal control framework that predicts the trajectories of surrounding vehicles using a Markov chain-based model. These predicted trajectories are then integrated into the collision avoidance constraints of the optimal control problem. Furthermore, we expound on a trajectory optimization technique under state constraints, utilizing a planner grounded in necessary condition and jump condition.

This planner demonstrates its capability to numerically resolve collision avoidance scenarios involving surrounding vehicles. The simulation results robustly validate the effectiveness of our proposed approach in adeptly handling interaction-based motion planning across diverse scenarios. The incorporation of the predictor and optimal trajectory planner mark significant advancements in ensuring collision-free and optimized trajectories for autonomous vehicles navigating highway environments.

\addtolength{\textheight}{-12cm}   





\begin{thebibliography}{99}

\bibitem{c1}
Wang, Weida, et al. “A multi-objective optimization energy management strategy for power split HEV based on velocity prediction." Energy 238 (2022): 121714.



\bibitem{c2}
Kim, Dongryul, Hung Duy Nguyen, and Kyoungseok Han. “State-Constrained Lane Change Trajectory Planning for Emergency Steering on Slippery Roads." IEEE Transactions on Vehicular Technology (2023).

\bibitem{c3}
Huangfu, Yigeng, et al. “An Improved Energy Management Strategy for Fuel Cell Hybrid Vehicles Based on Pontryagin's Minimum Principle." IEEE Transactions on Industry Applications 58.3 (2022): 4086-4097.

\bibitem{c4}
Singh, Amrik Singh Phuman, and Osamu Nishihara. “Trajectory tracking and integrated chassis control for obstacle avoidance with minimum jerk." IEEE Transactions on Intelligent Transportation Systems 23.5 (2021): 4625-4641.

\bibitem{c5}
Hu, Xiangwang, and Jian Sun. “Trajectory optimization of connected and autonomous vehicles at a multilane freeway merging area." Transportation Research Part C: Emerging Technologies 101 (2019): 111-125.

\bibitem{c6}
Ji, Kyoungtae, et al. “Hierarchical and game-theoretic decision-making for connected and automated vehicles in overtaking scenarios." Transportation research part C: emerging technologies 150 (2023): 104109.

\bibitem{c7}
Bersani, Mattia, et al. “An integrated algorithm for ego-vehicle and obstacles state estimation for autonomous driving." Robotics and Autonomous Systems 139 (2021): 103662.

\bibitem{c8}
Nguyen, Hung Duy, et al. “Linear Time-Varying MPC-based Autonomous Emergency Steering Control for Collision Avoidance." IEEE Transactions on Vehicular Technology (2023).

\bibitem{c9}
Nguyen, Hung Duy, Mooryong Choi, and Kyoungseok Han. “Risk-informed decision-making and control strategies for autonomous vehicles in emergency situations." Accident Analysis Prevention 193 (2023): 107305.

\bibitem{c10}
Jalalmaab, Mehdi, et al. “Model predictive path planning with time-varying safety constraints for highway autonomous driving." 2015 international conference on advanced robotics (icar). IEEE, 2015.

\bibitem{c11}
Nilsson, Julia, et al. “Predictive manoeuvre generation for automated driving." 16th International IEEE Conference on Intelligent Transportation Systems (ITSC 2013). IEEE, 2013.

\bibitem{c12}
Wang, Yijing, et al. “Trajectory planning and safety assessment of autonomous vehicles based on motion prediction and model predictive control." IEEE Transactions on Vehicular Technology 68.9 (2019): 8546-8556.

\bibitem{c13}
Kim, Namwook, Jongryeol Jeong, and Chunhua Zheng. “Adaptive energy management strategy for plug-in hybrid electric vehicles with Pontryagin’s minimum principle based on daily driving patterns." International Journal of Precision Engineering and Manufacturing-Green Technology 6 (2019): 539-548.

\bibitem{c14}
Locatelli, Arturo, and S. Sieniutycz. “Optimal control: An introduction." Appl. Mech. Rev. 55.3 (2002): B48-B49.



\bibitem{c15}
Raftery, Adrian E. “A model for high-order Markov chains." Journal of the Royal Statistical Society Series B: Statistical Methodology 47.3 (1985): 528-539.

\bibitem{c16}
Zhang, Han, Chang Liu, and Wanzhong Zhao. “Segmented Trajectory Planning Strategy for Active Collision Avoidance System." Green Energy and Intelligent Transportation 1.1 (2022): 100002.


\end{thebibliography}
\end{document}